\documentclass{article}

\usepackage{microtype}
\usepackage{graphicx}
\usepackage{subcaption}
\usepackage{booktabs} 
\usepackage{amsmath}

\usepackage{hyperref}

\usepackage[accepted]{mlsys2025}

\mlsystitlerunning{Pre-Attention Expert Prediction and Prefetching For MoE LLMs}

\begin{document}

\twocolumn[
\mlsystitle{Pre-Attention Expert Prediction and Prefetching for Mixture-of-Experts Large Language Models}




\begin{mlsysauthorlist}
\mlsysauthor{Shien Zhu}{to}
\mlsysauthor{Samuel Bohl}{to}
\mlsysauthor{Robin Oester}{to}
\mlsysauthor{Gustavo Alonso}{to}
\end{mlsysauthorlist}

\mlsysaffiliation{to}{Systems Group, D-INFK, ETH Zurich, Switzerland} 

\mlsyscorrespondingauthor{Shien Zhu}{shien.zhu@inf.ethz.ch}
\mlsyscorrespondingauthor{Gustavo Alonso}{alonso@inf.ethz.ch}

\mlsyskeywords{MoE, LLM, Efficient Inference, Expert Prefetching, Expert Prediction} 

\vskip 0.3in

\begin{abstract}
Mixture-of-Experts (MoE) Large Language Models (LLMs) efficiently scale-up the model while keeping relatively low inference cost. As MoE models only activate part of the experts, related work has proposed expert prediction and caching methods to prefetch the experts for faster inference. However, existing approaches utilize the activations from the previous layer for prediction, incurring low accuracy and leave the first layer unoptimized. Applying complex layers or even training standalone networks for better prediction introduces high computation overhead. 
In this paper, we propose pre-attention expert prediction to achieve accurate and lightweight expert prefetching. The key insight is that some functions in LLMs are ranking-preserving, indicating that matching the ranking of selected experts using simple linear functions is possible. Therefore, we utilize the activations before the attention block in the same layer with 2 linear functions and ranking-aware loss to achieve accurate prediction, which also supports prefetching in the first layer. Our lightweight, pre-attention expert routers achieve 93.03\% accuracy on DeepSeek V2 Lite, 94.69\% on Qwen3-30B, and 97.62\% on Phi-mini-MoE, showing about 15\% improvement on absolute accuracy over the state-of-the-art FATE results. 
\end{abstract}
]


\printAffiliationsAndNotice{}  

\section{Introduction}
The Mixture-of-Experts (MoE) architecture is widely adopted by the latest Large Language Models (LLMs) such as Llama 4 \cite{Llama_GitHub}, DeepSeek-V3 \cite{DeepSeekv3_ArXiv_2025}, and QWen-3 \cite{QWen3_ArXiv_2025}. MoE LLMs can achieve high cognition and generation performance while keeping inference cost relatively low. The key reason is that MoE LLMs only activate part of the experts in the Feed-Forward Networks (FFNs), which avoids the high computation cost of using all experts. For example, DeepSeek-V3 only activates 1 shared expert and 8 of 256 routed experts (activates 37 out of 671 billion parameters), significantly reducing runtime memory and computation overhead. As the MoE routers are placed inside the FFNs, the expert selection-loading-computation pipeline suffers from the long expert loading time, especially when the MoE models are too large to fit in the GPU memory.

To solve the expert-loading bottleneck in MoE LLMs, various prediction and caching methods have been proposed to efficiently prefetch and serve the experts. MoE-Infinity \cite{MoE-Infinity_ArXiv_2024} designs a sparsity-aware expert cache to trace the activated experts to reduce the Time-Per-Output-Token (TPOT). PopFetcher \cite{PopFetcher_ATC_2025} prefetches the experts of the next layer based on their popularity. HOBBIT \cite{HOBBIT_ArXiv_2024} proposes a multi-dimensional cache manager and dynamic expert loader to accelerate expert loading. Pre-Gated MoE \cite{Pre-Gate-MoE_ISCA_2024} modifies the gate function to select the experts of the next layer and applies caching methods for faster inference. 

However, we observe three problems in these prediction systems. First, it is very hard to achieve high prediction accuracy by predicting from the previous layer. SP-MoE \cite{SP-MoE_ArXiv_2025} prefetches the experts for speculative decoding that drafts multiple tokens per step and achieves more than 70\% of prediction accuracy in most layers. DuoServe-MoE \cite{DuoServe-MoE_ArXiv_2025} accelerates the inference by duo CUDA streams and a layer-level expert predictor with 54-67\% of top-2 accuracy. FATE \cite{FATE_ArXiv_2025} uses the activations from the previous layer for prediction, and the prediction part contributes 78.8\% to accuracy. In addition, many proposals report only the performance gain, without the prediction accuracy. Second, predicting from the previous layer has a limitation on prefetching for the first layer. Although they still predict the experts for the first layer, the actual prediction accuracy for the first and second layers is significantly lower than that of the other layers \cite{FATE_ArXiv_2025}. For this reason, AdapMoE \cite{AdapMoE_ICCAD_2024} tries to mitigate the prediction accuracy gap of the first few layers by integrating prefetching and cache management techniques. Third, increasing the complexity of the predictor can bring higher prediction accuracy, but also incur high computational overhead. The prediction latency has to be as short as possible to leave enough time to fetch the experts, otherwise the prefetching benefits will diminish.  

\begin{figure}[t]
\centering
\includegraphics[width=0.50\textwidth]{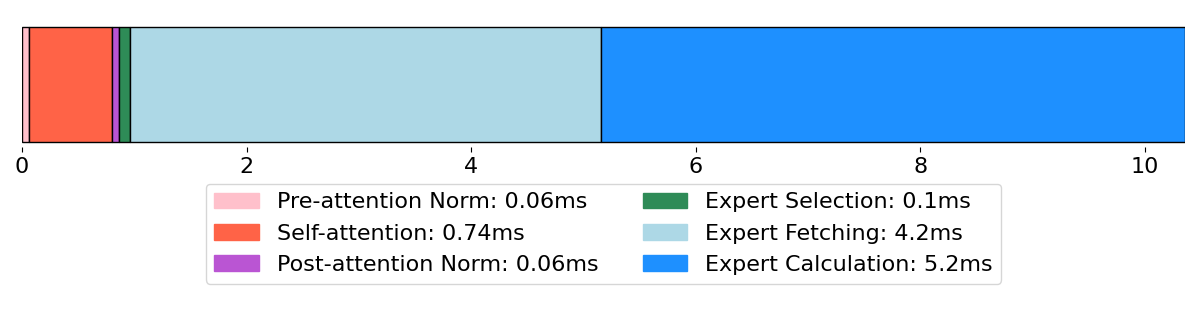}
\caption{Token generation pipeline in typical MoE architectures (profiled on DeepSeek-V2-Lite on a Nvidia V100 GPU).}
\label{fig:moe_pipeline}
\end{figure}

In this paper, we propose pre-attention expert prediction to achieve accurate and lightweight expert prefetching for MoE LLM inference. Our key insight is that the softmax and layer normalization functions are ranking-preserving, allowing ranking-based expert selection to be approximated by simple linear functions, as in the original expert router. First, in contrast to related work predicting based on the previous layer, we propose to predict experts within the same layer before the attention block to achieve a more accurate prediction. This also solves the problem of expert prefetching in the first layer. Second, we design lightweight pre-attention expert routers with 2 linear layers with the intermediate size being 2048, which has significantly lower computation cost than methods with standalone networks. Third, as we infer that matching the ranking of selected experts is possible, we propose ranking-aware loss functions to train the expert prediction routers. 

Our pre-attention expert routers achieve 93.03\% accuracy on DeepSeek V2 Lite, 94.69\% on Qwen3-30B, and 97.62\% on Phi-mini-MoE, showing 15\% improvement on absolute accuracy over the results in FATE \cite{FATE_ArXiv_2025}. 

\section{Background and Related Work}

\subsection{MoE Architecture and Inference Challenges}

Modern MoE architectures replace traditional FFNs in Transformers with collections of expert modules managed by learned routing functions~\cite{MoE_ArXiv_2017, Switch_JMLR_2022}. This paradigm has evolved from foundational models like GShard~\cite{gshard2020} and Switch Transformer~\cite{Switch_JMLR_2022} to recent large-scale implementations including DeepSeek-V2~\cite{DeepSeek-V2_ArXiv_2024}, DeepSeek-V3~\cite{deepseekv3}, Qwen3~\cite{QWen3_ArXiv_2025}, Phi-3~\cite{Phi-3_ArXiv_2024}, Mixtral 8x7B and 8x22B~\cite{mixtral2023}, DBRX~\cite{dbrx2024}, Hunyuan-Large~\cite{hunyuan2024}, and many other models. These models demonstrate the scalability of MoE architectures, with total parameters ranging from Mixtral 8x7B's 47 billion to DeepSeek-V3's 671 billion parameters, while maintaining sparse activation patterns that significantly reduce computational requirements during inference.

\begin{figure}[t]
\centering
\includegraphics[width=0.48\textwidth]{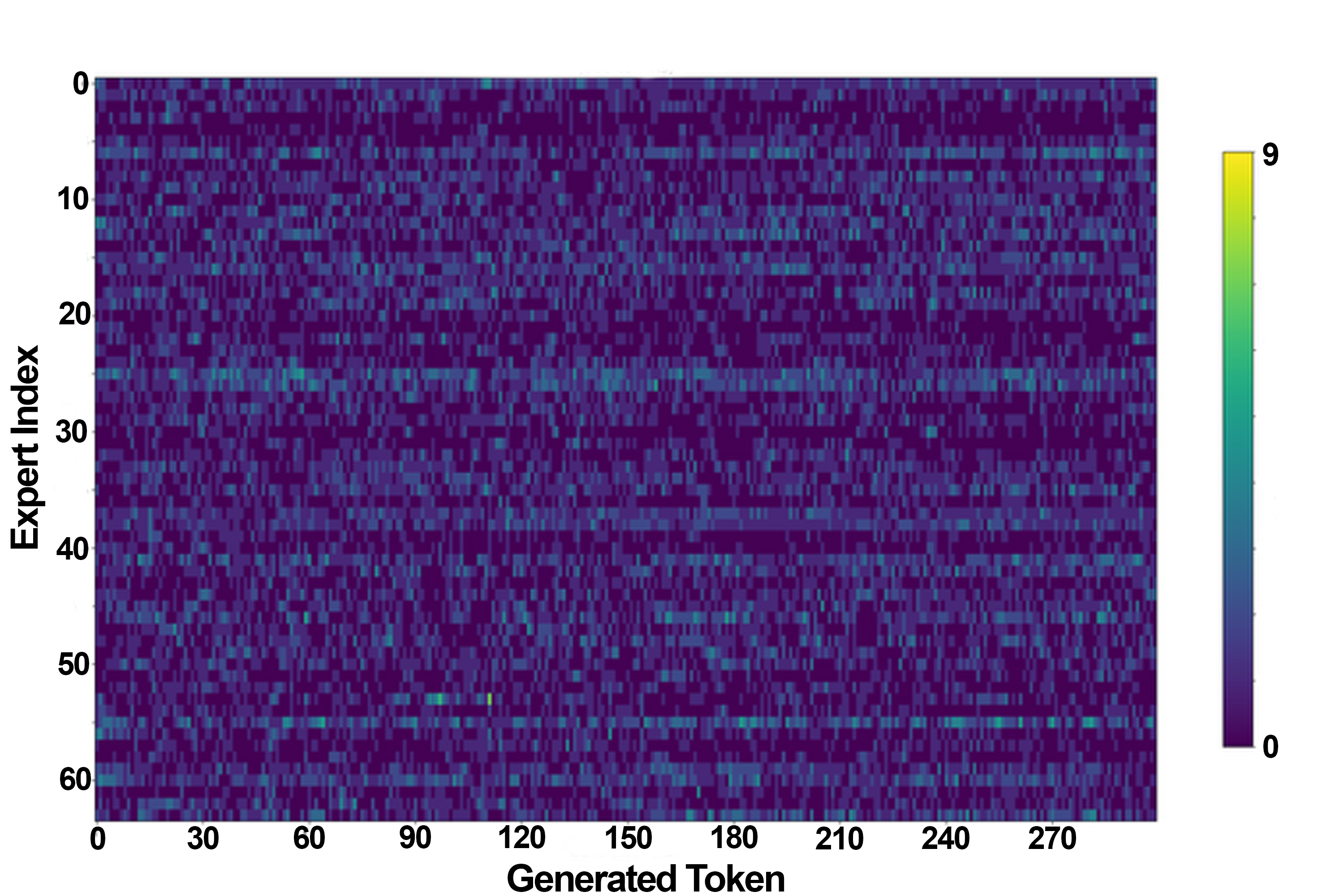}
\hfill
\includegraphics[width=0.48\textwidth]{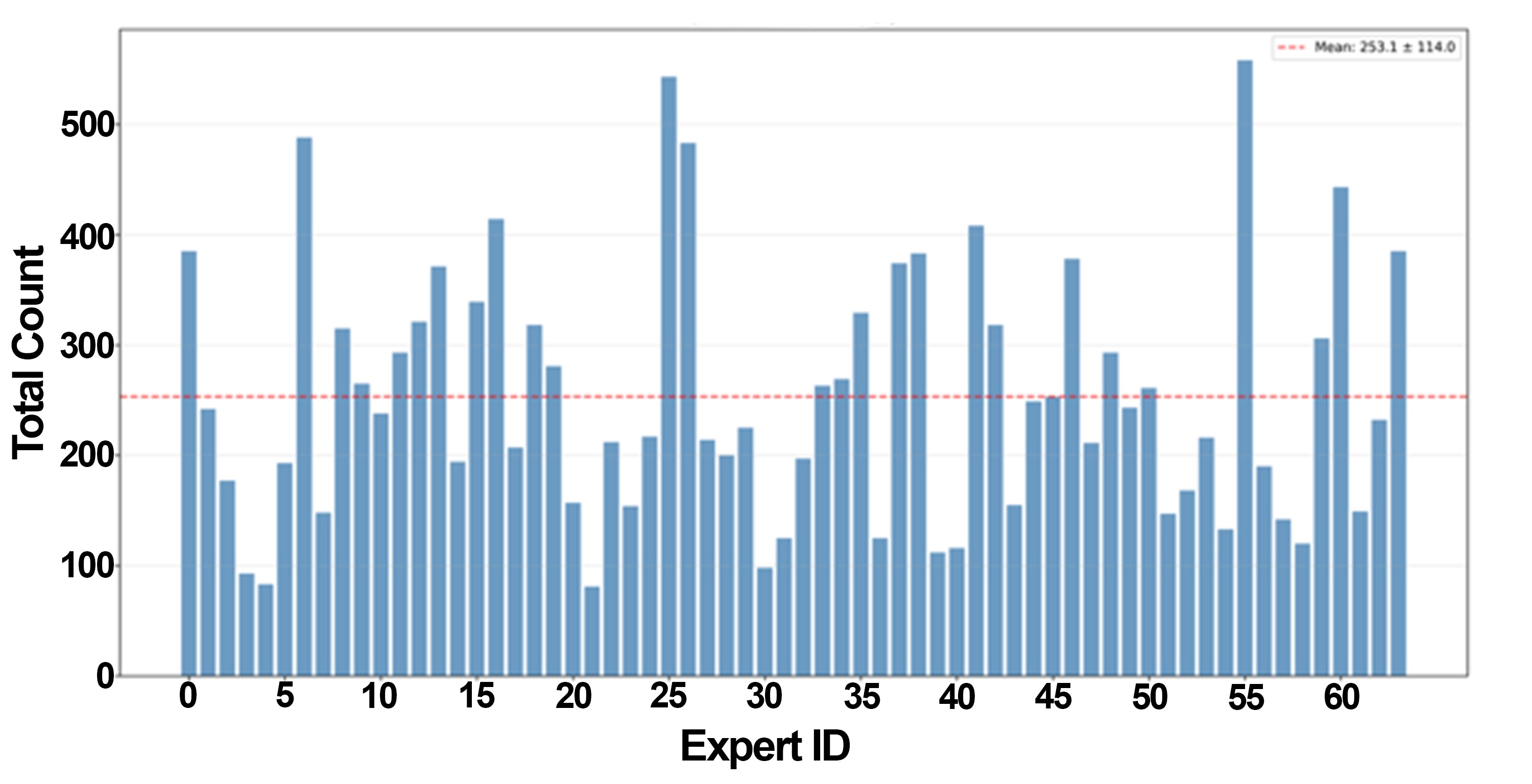}
\caption{(a) Aggregated expert invocation heatmap and (b) Distribution of expert activation frequencies on DeepSeek-V2-Lite across the first 9 layers when generating 300 tokens.} 
\label{fig:expert_activation_analysis}
\end{figure}

During inference, each input token is dynamically assigned to a subset of experts based on the router's decisions, typically selecting the top-k experts with highest routing scores. For instance, Hunyuan-Large activates 52 billion out of 389 billion parameters~\cite{hunyuan2024}. Although this sparse activation pattern enables significant computational savings compared to dense alternatives such as GPT-3~\cite{glam2021}, it introduces substantial challenges for practical deployment in resource-constrained environments. As Figure \ref{fig:moe_pipeline} shows, fetching the experts becomes the bottleneck. 
The issue arises from the unpredictable nature of expert routing decisions. Unlike dense models where all parameters are accessed, MoE models dynamically select experts depending on the current input. This unpredictability is evident in our analysis of DeepSeek-V2-Lite (Figure~\ref{fig:expert_activation_analysis}). The expert activation patterns across the first 9 layers during 300 tokens of inference exhibit a high Shannon entropy of 0.976, confirming the near-uniform expert distribution and the inherent challenge of anticipating which experts will be activated for any given token. 

This unpredictability creates I/O bottlenecks when expert must be loaded into GPU memory during inference. Critical challenges include: dynamic load imbalance where expert utilization becomes severely skewed~\cite{auxloss2024,prototyperouting2024}, communication overhead from inter-GPU all-to-all patterns required for distributed expert routing~\cite{lina2022,c2r2024}, and memory efficiency bottlenecks during expert caching and dynamic loading~\cite{expertflow2024,moetuner2024}.
In addition, recent MoE architectures have introduced additional complexity through shared and fine-grained expert designs~\cite{sun2024}. DeepSeekMoE subdivides experts into smaller, specialized units while maintaining shared experts that are always activated. This design improves expert specialization but further complicates prediction due to the increased number of routing decisions per layer. Similarly, auxiliary-loss-free load balancing strategies~\cite{auxloss2024} and dynamic expert selection mechanisms~\cite{hardertasks2024} aim to address routing instabilities but create more complex prediction scenarios for inference optimization systems.

\subsection{Expert Prefetching Methods}

\subsubsection{Caching-Based Methods}
Expert caches are specialized data caches combining traditional software cache-related techniques and optimizations for MoE LLM expert serving. MoE-Infinity \cite{MoE-Infinity_ArXiv_2024} is a sparsity-aware cache exploiting the skewed reuse patterns of MoE LLM experts. Their method traces the sparse set of activated experts to guide expert prediction, improving inference efficiency on personal machines. HOBBIT \cite{HOBBIT_ArXiv_2024} is a caching-based method that replaces less critical cache-miss experts with low-precision ones to reduce the loading latency. They efficiently manage the expert cache and achieve good speedup in decoding inference, but the cache hit rate is only around 55\% on Mixtral-8x7B and Phi-MoE.

\subsubsection{Prediction-Based Methods}
Existing works mainly focus on cross-layer expert prediction and prefetching, as illustrated in Fig.\ref{fig:moe-layer-predictors} (b). PopFetcher \cite{PopFetcher_ATC_2025} aims to accelerate MoE LLM training by utilizing the communication bandwidth when computing the attention blocks to hide the expert loading latency. They prefetch the experts for the next layer based on the expert popularity by a heuristic approach considering the skewed and correlated expert selection patterns.
DuoServe-MoE \cite{DuoServe-MoE_ArXiv_2025} applies two CUDA streams to overlap the expert loading and computation in the prefilling stage. Their lightweight layer-level expert predictor achieves 54-67\% of top-2 accuracy and 90.3-95.5\% hit-1 accuracy on Mixtral-8x7B and Mixtral-8x22B. SP-MoE \cite{SP-MoE_ArXiv_2025} optimizes the expert prefetching for speculative decoding that drafts multiple tokens per step. They achieve about 70-90\% of prediction accuracy in most layers thanks to the similarity of draft and target methods and about 44.3\% of hit rate on the drafted tokens. Note that SP-MoE targets a different workload from this paper.

\subsubsection{Hybird Caching- and Prediction-Based Methods}
Combining the caching with prediction may further boost the expert prediction and serving accuracy. FATE \cite{FATE_ArXiv_2025} applies prediction, caching, and mixed-precision experts in the perfetching system. It achieves 78.8\% accuracy by prediction and 97.2\% accuracy by loading all experts above a confidence score threshold. Pre-Gated MoE \cite{Pre-Gate-MoE_ISCA_2024} modifies the expert router to select the experts for the next layer instead of the current layer. They also enhance the inference system with caching to efficiently prefetch experts, reducing GPU memory usage and hiding expert-loading latency. 

\section{Motivation}
\begin{figure}[t]
\centering
\includegraphics[width=0.48\textwidth]{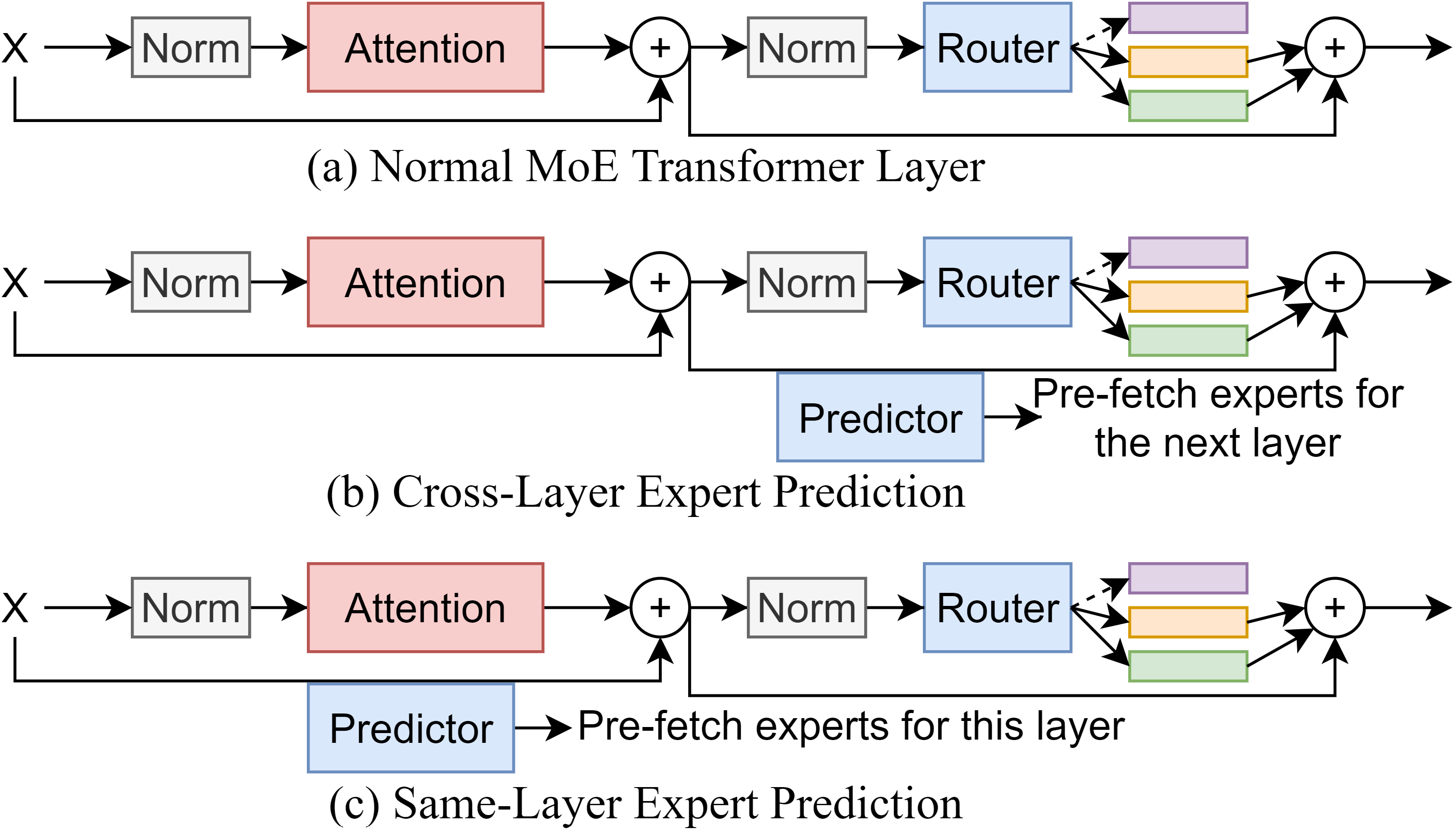}
\caption{Example MoE layers with and without expert prediction.}
\label{fig:moe-layer-predictors}
\end{figure}



While existing cross-layer prediction approaches have demonstrated considerable success~\cite{FATE_ArXiv_2025,MoE-Infinity_ArXiv_2024}, they face inherent architectural complexity and cross-layer dependency management challenges. As Fig.\ref{fig:moe-layer-predictors} (c) shows, same-layer prediction using pre-attention normalized weights offers a fundamentally different approach that leverages temporal proximity between token representation formation and expert selection decisions.

\subsection{Hypothesis for Same-Layer Prediction}
We hypothesize that same-layer expert prediction may be feasible based on two key observations. First, pre-attention weights contain more recent information than previous layer outputs, capturing token representation at the critical juncture immediately before expert routing decisions. This temporal proximity suggests potential for more accurate routing decisions compared to cross-layer extrapolation.

Second, expert selection fundamentally requires ranking rather than exact continuous value prediction~\cite{MoE_ArXiv_2017,Switch_JMLR_2022}. As softmax and layer normalization are ranking-preserving, recent work suggests that attention mechanisms may exhibit approximately ranking-preserving properties~\cite{nguyen2025,yang2025}. If this property holds, linear functions with a simple non-linear activation function may potentially capture the ranking relationships necessary for expert selection.  

\subsection{Benefits of Same-Layer Prediction}
Same-layer prediction has several practical advantages over cross-layer approaches. \textbf{First}, it eliminates the bootstrap problem for initial MoE layers where no previous gate input exists, as every layer has its predictor. 
\textbf{Second}, same-layer prediction eliminates architectural complexity by removing cross-layer communication overhead. Unlike methods that require coordinating information across transformer layers, same-layer approaches operate independently using readily available pre-attention weights, eliminating the need for additional memory buffers, state management across layers, or inter-layer communication protocols.
\textbf{Third}, same-layer prediction offers improved interpretability through direct relationships between token characteristics and expert selection within the same computational context. This enables more straightforward analysis of prediction behavior compared to cross-layer approaches that must account for complex inter-layer dependencies.
\textbf{Fourth}, pre-attention prediction enables parallel execution with self-attention processing, which requires $0.74\text{-}1.13$ milliseconds across different hardware configurations. This parallel execution window provides sufficient time for both prediction computation and expert prefetching without introducing additional latency to the inference pipeline.

\section{Pre-Attention Expert Prediction}

\subsection{Problem Formulation}

We formulate the general expert prediction problem for different MoE configurations and deployment scenarios. For example, Qwen3-30B selects 8 experts, but Phi-mini selects 2 experts in one FFN. Thus, we predict the top-k expert indices that will be selected by the original routers.

Let $\mathbf{X} \in {R}^d$ denote the pre-attention weights for a given token at layer $l$, where $d$ represents the hidden dimension. The standard MoE routing mechanism computes expert selection through equations (\ref{eq-MoE-g})-(\ref{eq-MoE-Y}), where $\mathbf{W}_g \in {R}^{E \times d}$ are the gating parameters for $E$ experts, and $\text{TopK}(\cdot, k)$ selects the indices of the $k$ highest-scoring experts.
\begin{align}
\mathbf{g} &= \text{Softmax}(\mathbf{W}_g \cdot \mathbf{X}) 
\label{eq-MoE-g} \\
\hat{\mathbf{Y}}_{\text{true}} &= \text{TopK}(\mathbf{g}, k)
\label{eq-MoE-Y}
\end{align}
Our objective is to learn a mapping $f(\mathbf{X}; \theta) \rightarrow \hat{\mathbf{Y}}$ that accurately predicts the expert selection $\hat{\mathbf{Y}} = \{y_1, y_2, \ldots, y_k\}$ using only the pre-attention information available within the current layer. As equations (\ref{eq-MoE-new-s})-(\ref{eq-MoE-new-Y}) show, $\mathbf{s} \in {R}^E$ represents the predicted expert selection scores, and we select the top-k experts by ranking these scores directly.
\begin{align}
\mathbf{s} &= f(\mathbf{X}; \ \theta) 
\label{eq-MoE-new-s} \\
\hat{\mathbf{Y}} &= \text{TopK}(\mathbf{s}, k)
\label{eq-MoE-new-Y} 
\end{align}
We address three distinct deployment scenarios through different formulations. For standard deployment scenarios, we predict the precise set of k experts that will be selected. For over-provisioning scenarios, we predict a larger set of experts (e.g., 10 instead of 6 for DeepSeek V2 Lite) to achieve higher hit rates at the cost of increased I/O overhead. For I/O bandwidth-constrained edge scenarios where only one expert can be loaded in parallel with attention computation, we evaluate top-1 accuracy, which measures whether the single highest-scoring predicted expert is among the k experts that will actually be selected by the routing function.

\begin{table}[!b]
\centering
\small
\begin{tabular}{lc}
\toprule
\textbf{Hardware Configuration} & \textbf{Timing (ms)} \\
\midrule
\multicolumn{2}{l}{\textit{Pre-Attention Norm}} \\
Tesla V100-SXM2-32GB & 0.1292 $\pm$ 0.0120 \\
NVIDIA A100-PCIE-40GB & 0.0771 $\pm$ 0.0007 \\
NVIDIA A100 80GB PCIe & 0.0750 $\pm$ 0.0015 \\
\midrule
\multicolumn{2}{l}{\textit{Self-Attention}} \\
Tesla V100-SXM2-32GB & 1.1279 $\pm$ 0.0388 \\
NVIDIA A100-PCIE-40GB & 0.7607 $\pm$ 0.0069 \\
NVIDIA A100 80GB PCIe & 0.7385 $\pm$ 0.0074 \\
\midrule
\multicolumn{2}{l}{\textit{Post-Attention Norm}} \\
Tesla V100-SXM2-32GB & 0.1292 $\pm$ 0.0120 \\
NVIDIA A100-PCIE-40GB & 0.0823 $\pm$ 0.0012 \\
NVIDIA A100 80GB PCIe & 0.0797 $\pm$ 0.0040 \\
\midrule
\multicolumn{2}{l}{\textit{Expert Selection}} \\
Tesla V100-SXM2-32GB & 0.1432 $\pm$ 0.0138 \\
NVIDIA A100-PCIE-40GB & 0.0972 $\pm$ 0.0025 \\
NVIDIA A100 80GB PCIe & 0.1018 $\pm$ 0.0039 \\
\midrule
\multicolumn{2}{l}{\textit{Expert Computation}} \\
Tesla V100-SXM2-32GB & 10.3075 $\pm$ 1.7038 \\
NVIDIA A100-PCIE-40GB & 6.1970 $\pm$ 1.0668 \\
NVIDIA A100 80GB PCIe & 6.8111 $\pm$ 1.1864 \\
\bottomrule
\end{tabular}
\caption{Timing comparison of key transformer operations in DeepSeek-V2-Lite across different GPU configurations. All measurements represent mean $\pm$ standard deviation over 50 samples.}
\label{tab:timing_analysis}
\end{table}

\subsection{Pre-Attention Prediction Workflow}

Our approach exploits the natural pipeline timing of transformer inference to perform expert prediction with minimal overhead. During the standard transformer forward pass, pre-attention normalization produces weights that capture the token representation immediately before expert routing. We clone these weights to the CPU for parallel prediction computation while the GPU continues with self-attention. 

The prediction process operates in three stages. First, we extract the pre-attention weights $\mathbf{X}$ immediately after layer normalization and before self-attention computation. Second, we feed these weights through our trained prediction model $f_l(\mathbf{X}; \theta_l)$ to generate expert probability scores, where we maintain a separate predictor for each layer $l$ with layer-specific parameters $\theta_l$. This layer-wise approach allows each predictor to specialize in the unique expert selection patterns characteristic of its corresponding transformer layer. Third, we select the top-k experts based on these scores for prefetching or caching decisions.
The critical advantage of this approach is timing. The prediction computation occurs in parallel with self-attention, which typically requires $0.73\text{-}1.13$ milliseconds across different hardware configurations (Table~\ref{tab:timing_analysis}). This parallel execution window provides sufficient time for both prediction computation ($0.075\text{-}0.129$ ms) and an early start for expert prefetching operations without introducing additional latency to the inference pipeline. 

\begin{figure}[t]
\centering
\includegraphics[width=0.4\textwidth]{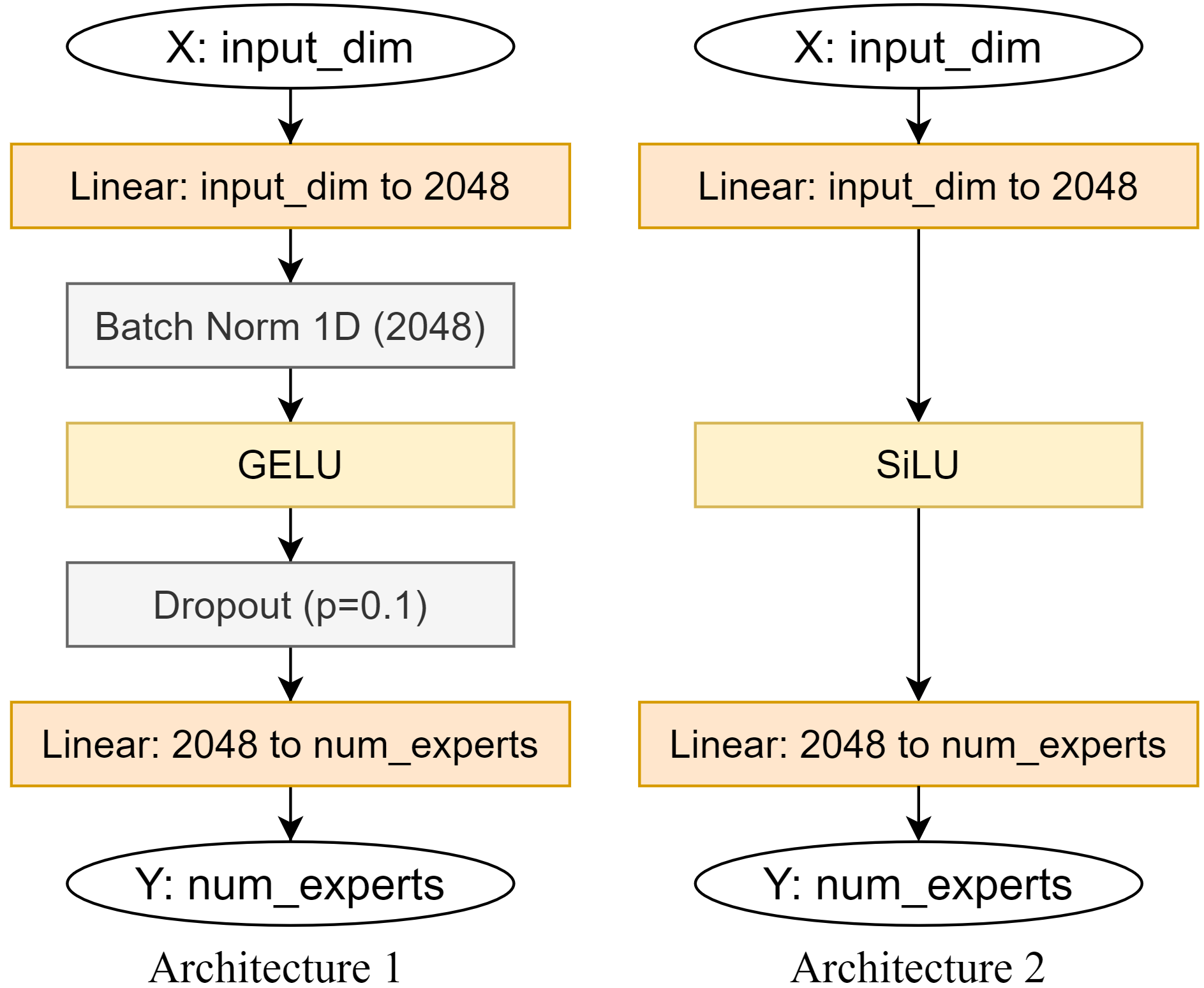}
\caption{Expert Selector Architecture Comparison}
\label{fig:predictor_architecture}
\end{figure}

\subsection{Predictor Architectures}
We formulate expert prediction as a multi-label classification problem for training purposes, where we create binary labels indicating whether each expert was among the top-k selected experts in the ground truth. Afterwards, during inference, we select experts by ranking the predicted confidence score of all experts and choosing the top-k highest scoring experts, similar to the original MoE gating mechanism.

We evaluated two different architectures for expert selection, as illustrated in Figure~\ref{fig:predictor_architecture}. Architecture 1 employs linear layers and additional regularization components: a linear layer mapping the input dimension to a hidden dimension (2048), followed by BatchNorm1d, GELU activation, dropout (p=0.1), and a final linear layer that outputs logits for each expert. Architecture 2 uses a streamlined design that contains two linear layers and a SiLU activation in between. This configuration can evaluate the impact of different regularization techniques and activation functions on expert selection accuracy across different MoE models.

\subsection{Training Dataset and Representative MoE Models}
We obtain the training datasets for the predictors from MMLU~\cite{MMLU_ICLR_2021} to ensure various realistic expert usage patterns. We collect $10\text{M}$ training samples comprising pre-attention activations and corresponding ground-truth affinity scores and expert selections from three actual MoE models.

We selected three representative MoE models that span different scales and architectural configurations while fitting within our hardware constraints. DeepSeek V2 Lite \cite{DeepSeek-V2_ArXiv_2024} enables direct comparison with existing research including MoE-Infinity \cite{MoE-Infinity_ArXiv_2024} and FATE \cite{FATE_ArXiv_2025}, Qwen3-30B-A3B \cite{QWen3_ArXiv_2025} tests scalability to larger architectures, and Phi-mini-MoE \cite{Phi-3_ArXiv_2024} validates effectiveness in resource-constrained scenarios. Table~\ref{tab:moe_architectures} summarizes the key architectural parameters of these models. The varying expert counts (16, 64, 128), activation patterns (2, 6, 8), and model scales (7.6B, 16.4B, 30.5B) collectively demonstrate the generalizability of our approach across the spectrum of practical MoE deployments. Their activation ratios (6.3-12.5\%) provide a challenging prediction scenario that forecasts expert selection from a large candidate pool.

\begin{table}[t]
\centering
\scriptsize
\begin{tabular}{lccc}
\toprule
\textbf{Parameter} & \textbf{DeepSeek-V2-Lite} & \textbf{Qwen3-30B} & \textbf{Phi-mini-MoE} \\
\midrule
Total Params & 16.4B & 30.5B & 7.6B \\
Active Params & 2.4B & 3.3B & 2.4B \\
Layers & 27 & 48 & 32 \\
Experts/Layer & 64 & 128 & 16 \\
Active Experts & 6 (+ 2 shared) & 8 & 2 \\
Activation Ratio & 9.4\% & 6.3\% & 12.5\% \\
\bottomrule
\end{tabular}
\caption{Representative MoE models used in evaluation.}
\label{tab:moe_architectures}
\end{table}

\subsection{Loss Function Design}
The loss function is critical for expert prediction, as the problem exhibits unique characteristics from standard multi-label classification tasks. We systematically evaluate multiple loss functions to identify an effective training objective for our pre-attention prediction approach. We find that the loss function with multi-label classification principles yields superior results than regression-based approaches. Rather than regression-based approximating the continuous routing affinity scores directly, we find that the classification-oriented loss function based on discrete expert selections is more effective, achieving $90.72\%$ vs $86.61\%$ accuracy in comparative experiments.

Therefore, we incorporate several optimizations inspired by recent advances in multi-label learning. We use focal and weighted binary cross-entropy loss to address class imbalance inherent in the multi-label setting, where only top-k experts are positive for each sample. In addition, we incorporate ranking-aware loss components to preserve the relative order of experts. 

\subsubsection{Affinity Score Distribution}

\begin{figure}[!t]
\centering
\includegraphics[width=0.48\textwidth]{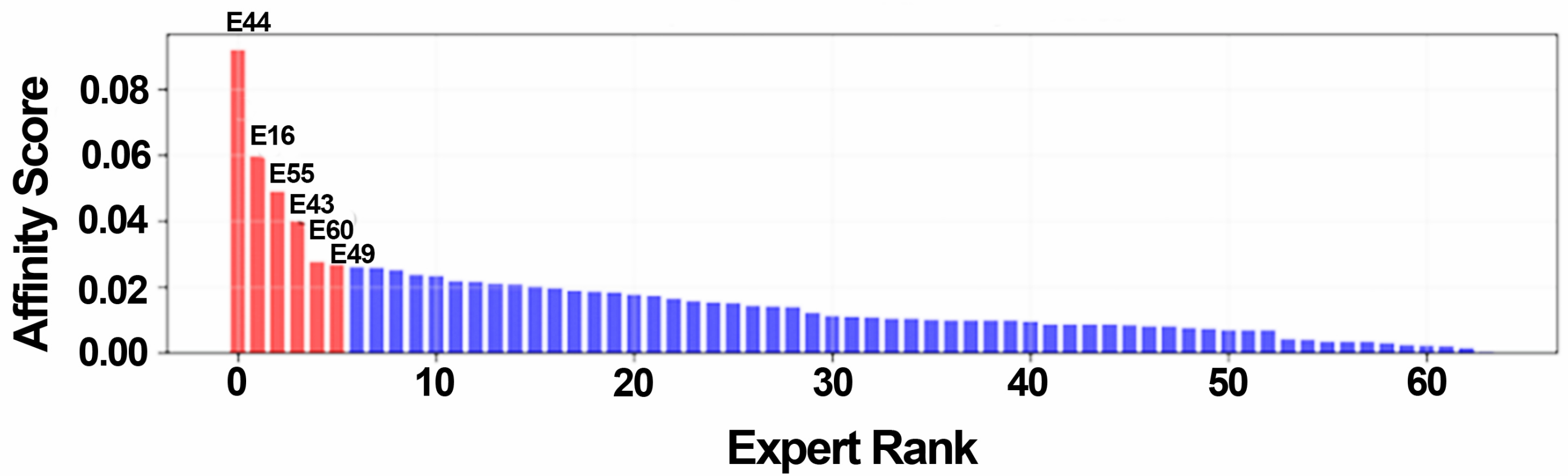}
\caption{Sample of sorted true affinity scores.}
\label{fig-score-sample}
\end{figure}

We observe an interesting pattern when inspecting the affinity scores produced by the MoE gating mechanism. As Fig. \ref{fig-score-sample} sampled on a DeepSeek-V2-Lite layer shows, the distribution of expert scores exhibits a distinct three-tier structure: the top-ranked experts (typically the first few ones) show significantly high affinity scores, followed by a middle tier where scores remain relatively flat across a large number of experts, and finally a sharp drop-off for the lowest-ranked experts. As only a few experts receive high routing confidence, the expert prediction in the moderate region (experts 5 and 6 in Fig. \ref{fig-score-sample}) is very challenging. 
Leveraging this observation, we design the loss function to reflect the inherent ranking structure of expert selection. Rather than treating all experts equally, we assign higher importance to top-ranked experts, moderate importance to middle-tier experts, and lower weight to remaining experts.

\subsubsection{Loss Function Formulations}

We evaluated three primary categories of loss functions through systematic single batch experimentation in our training dataset of $10\text{M}$ MMLU-derived samples.

\textbf{Regression-based approaches} treat expert prediction as a continuous score prediction problem. We implement a Mean Squared Error (MSE) loss that directly predicts the routing scores produced by the actual MoE gating mechanism:
\begin{align}
\mathcal{L}_{\text{MSE}}(\theta) = \frac{1}{N}\sum_{i=1}^{N} \sum_{j=1}^{E} (s_{ij} - \hat{s}_{ij})^2
\end{align}
where $s_{ij}$ represents the ground truth routing score for expert $j$ on sample $i$, and $\hat{s}_{ij}$ denotes the predicted score. While conceptually straightforward, this approach achieves only $86.61\%$ top-k accuracy under standardized evaluation conditions, indicating that direct score regression fails to capture the discrete nature of expert selection decisions.

\textbf{Standard multi-label classification} formulates the problem as binary expert selection prediction using weighted binary cross-entropy loss that prioritizes top-ranked experts. As equations (\ref{eq-wbce})-(\ref{eq-wbce-end}) show, the loss function depends on two conditions: if expert $j$ is among the top-k selected experts for sample $i$ and if expert $j$ is a top-10 expert in groundtruth for sample $i$. $\hat{z}_{ij}$ represents the predicted logits and $\sigma(\cdot)$ is the sigmoid function. This weighted formulation achieves 90.19\% accuracy with Architecture 2, outperforming the regression-based approach for expert selection.
\begin{align}
\mathcal{L}_{\text{WBCE}}(\theta) &= -\frac{1}{N \cdot E}\sum_{i=1}^{N} \sum_{j=1}^{E} w_{ij} l_{ij} 
\label{eq-wbce}\\
l_{ij} &=  \begin{cases} 
\log(\sigma(\hat{z}_{ij})) & \text{if expert } j \text{ is top-k} \\
\log(1-\sigma(\hat{z}_{ij}))  & \text{otherwise}
\end{cases}
\\
w_{ij} &= \begin{cases} 
3.0 & \text{if expert } j \text{ is real top-10} \\
0.5 & \text{otherwise}
\end{cases} 
\label{eq-wbce-end}
\end{align}

\begin{table}[!t]
\centering
\begin{tabular}{lcc}
\toprule
\textbf{Loss Function} & \textbf{Architecture 1} & \textbf{Architecture 2} \\
\midrule
MSE Regression & $86.61\%$ & $86.61\%$ \\ 
Weighted BCE  & $89.00\%$ & $\mathbf{90.19\%}$ \\ 
Focal Loss & $86.40\%$ & $87.64\%$ \\ 
Ranking-aware BCE & $\mathbf{89.19}\%$ & $89.95\%$ \\  
\bottomrule
\end{tabular}
\caption{Single-Epoch Loss Function Performance Comparison}
\label{tab:loss_comparison}
\end{table}

\textbf{Ranking-aware multi-label classification} incorporates the inherent ranking structure of expert selection through weighted and ranking-aware loss components. As equations (\ref{eq-ranking})-(\ref{eq-ranking-end}) show, the $\mathcal{L}_{\text{WBCE}}(\theta)$ part is almost the same as the previous loss function, except that the weighted binary cross-entropy component assigns greater importance to top-ranked experts ($w_{ij} = 1.5$ for experts ranking 11-30).
\begin{align}
\mathcal{L}_{\text{total}}(\theta) & = \mathcal{L}_{\text{WBCE}}(\theta) + \lambda \mathcal{L}_{\text{ranking}}(\theta) 
\label{eq-ranking} \\
w_{ij} &= \begin{cases} 
3.0 & \text{if expert } j \text{ is real top-10} \\
1.5 & \text{if expert } j \text{ is real top-11 to 30} \\
0.5 & \text{otherwise}
\end{cases} 
\\
\mathcal{L}_{\text{ranking}}(\theta) &= \sum_{i=1}^{N} \sum_{j,k \in \text{top-10}}^{s_{ij} > s_{ik}} \text{ReLU}(m - (s^{raw}_{ij} - s^{raw}_{ik})) 
\label{eq-ranking-end}
\end{align}
The pairwise ranking loss $\mathcal{L}_{\text{ranking}}(\theta)$ ensures relative ordering preservation within highly-ranked experts,
where only valid expert pairs ($s_{ij} > s_{ik}$) contribute to the loss. The margin $m$ is set to $[0.1]$ according to the validation performance. Our formulation combines weighted binary cross-entropy with pairwise ranking constraints, achieving 89.19\% accuracy with Architecture 1 and representing the best-performing approach for that architecture. Though the ranking-aware loss function does not achieve the highest accuracy on single-epoch experiment for architecture 2, it shows obvious better accuracy compared with focal loss~\cite{lin2018focal} and MSE regression.


\subsubsection{Experimental Loss Function Comparison}
\label{section:loss_comparison}

We evaluate the loss function variants using standardized single-epoch training on $10\text{M}$ samples to ensure fair comparison. Table~\ref{tab:loss_comparison} presents the comparative results across different formulations and architectures. The ranking-aware BCE and weighted BCE formulation achieve the highest accuracy of $90.19\%$ and $89.19\%$ respectively. The weighted component addresses the class imbalance, while the ranking component ensures that high-importance expert relationships are preserved during training.

Focal loss~\cite{lin2018focal}, designed for extreme class imbalance, achieves only $86.40\%$ accuracy in our experiments. This lower performance suggests that the class imbalance in expert prediction is better addressed through explicit weighting rather than dynamic loss scaling, likely due to the structured nature of expert selection patterns.

Based on the loss function analysis presented in Table~\ref{tab:loss_comparison}, we employ ranking-aware BCE loss for Architecture 1 and weighted BCE loss for Architecture 2, selecting the best formulation for each architecture for higher accuracy.

\begin{figure}[t]
\centering
\includegraphics[width=0.48\textwidth]{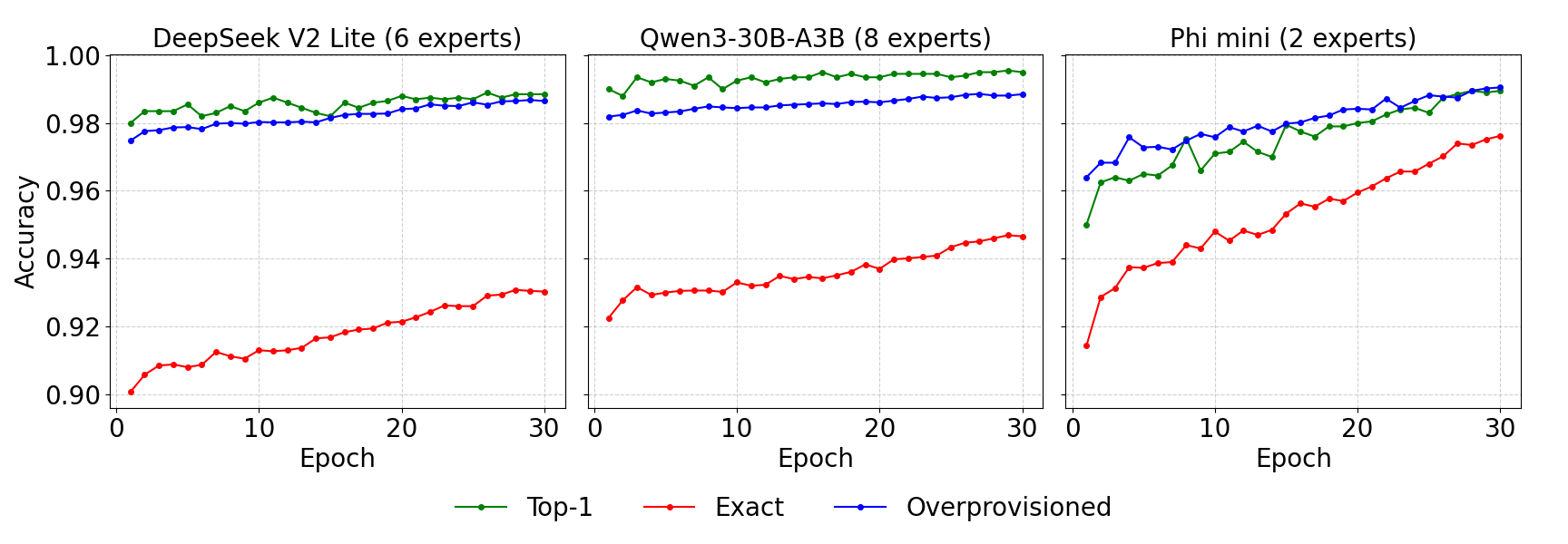}
\caption{Expert Prediction Accuracy Architecture 1}
\label{fig:expert_predictions_30ep}
\end{figure}

\subsubsection{Hyperparameter Selection}

We determine near-optimal hyper-parameters for our ranking-aware loss through grid search on validation data. The ranking loss weight $\lambda = 0.3$ in equation (\ref{eq-ranking}) provides an excellent balance between classification accuracy and ranking preservation. Lower values ($\lambda < 0.1$) reduce ranking quality while higher values ($\lambda > 0.5$) hurt overall classification performance.

The margin parameter $m$ in the ranking loss is set to $0.1$ based on the typical separation between expert routing scores in our training data. Larger margins ($> 0.2$) make the ranking constraints too strict, while smaller margins ($< 0.05$) provide insufficient ranking signal during training.

\section{Experimental Results}

\subsection{Experimental Setup}
We evaluate our pre-attention expert prediction across three representative MoE models: DeepSeek V2 Lite~\cite{DeepSeek-V2_ArXiv_2024}, Qwen3~\cite{QWen3_ArXiv_2025}, and Phi-mini~\cite{Phi-3_ArXiv_2024}, spanning different architectural configurations and expert selection strategies. Our training regimen employs $10\text{M}$ samples in $30$ epochs to achieve high accuracy across these models. We conducted all experiments on a system equipped with an NVIDIA TITAN RTX 24GB GPU, 128GB of system memory, and 8 CPU cores. Training the predictors for each Transformer layer using GPU is necessary, given the large sample volume. Nevertheless, GPU is not a strict requirement for inference of the predictors. CPU-only inference of the predictors is sufficient for deployment scenarios where the predictor overhead must be minimized.

We evaluate three metrics. First, the exact match accuracy, where the prediction must precisely identify the selected experts. Exact match accuracy measures the percentage of predictions that correctly identify all k selected experts. Second, the over-provisioning accuracy, where loading additional experts can achieve higher hit rates at an increased I/O cost. Third, the top-1 accuracy of the predictor, which evaluates the percentage of cases where the single highest-scoring predicted expert is among the k experts that will actually be selected. This is critical for edge deployment scenarios where I/O bandwidth constraints limit parallel expert loading to a single expert during attention computation. 

\subsection{Prediction Accuracy Results}

Our experimental results demonstrate substantial improvements over existing approaches across all evaluated models. As Table \ref{tab:final-prediction} shows, we achieve $93.03\%$ exact-match accuracy on DeepSeek-V2-Lite, representing a $15\%$ improvement on absolute accuracy over Fate's~\cite{FATE_ArXiv_2025} $78.79\%$ decoding accuracy. Qwen3-30B achieves approximately $94.69\%$ exact-match accuracy, while Phi-mini reaches $97.62\%$ accuracy, demonstrating that simpler MoE configurations benefit more from our approach.

\begin{figure}[t]
\centering
\includegraphics[width=0.48\textwidth]{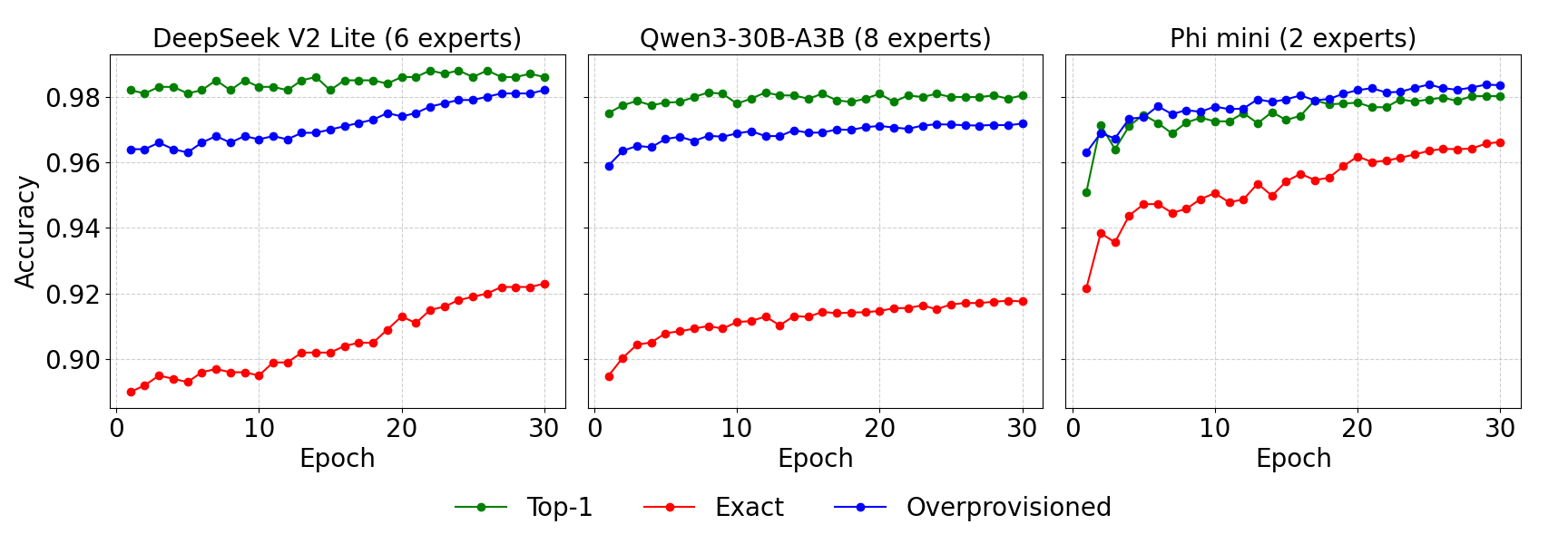}
\caption{Expert Prediction Accuracy Architecture 2}
\label{fig:expert_predictions_30ep_arch2}
\end{figure}

The superior performance on Phi-mini could highlight an important characteristic of our method: prediction accuracy might correlate with model complexity. Simpler routing decisions are easier to predict using pre-attention weights, while more complex routing patterns in larger models present greater challenges but still achieve substantial improvements over existing methods.

The over-provisioning results reveal the practical trade-offs available for different deployment scenarios. Loading 10 experts instead of the required 6 for DeepSeek achieves $98.65\%$ hit rate, representing around $67\%$ additional I/O overhead for a $4.5\%$ improvement in hit rate. Similarly, over-provisioning Qwen3 with 12 experts loaded instead of 8 achieves $98.81\%$ accuracy, while Phi-mini with 3 experts instead of the required 2 leads to $99.05\%$ accuracy. These trade-offs prove attractive for cloud deployments where I/O bandwidth is abundant, but prediction accuracy is critical for performance.

Top-1 accuracy results demonstrate the effectiveness of our approach for I/O bandwidth-constrained edge scenarios. We achieve $98.85\%$ top-1 accuracy on DeepSeek V2 Lite, $99.55\%$ on Qwen3-30B, and $98.95\%$ on Phi-mini-MoE. These results indicate that our prediction method can reliably identify at least one correct expert for parallel loading during attention computation, providing substantial benefits for edge deployments where I/O bandwidth constraints limit the number of experts that can be loaded simultaneously with layer execution.

\begin{table}[t]
\centering
\scriptsize
\begin{tabular}{llccc}
\toprule
\textbf{Arch} & \textbf{Accuracy} & \textbf{DeepSeek-V2-Lite} & \textbf{Qwen3-30B} & \textbf{Phi-mini-MoE} \\
\midrule
  & Exact-match       & \textbf{93.03\%} & \textbf{94.69\%} & \textbf{97.62\%} \\
1 & Over-provis.      & \textbf{98.65\%} & \textbf{98.81\%} & \textbf{99.05\%} \\
  & Top-1             & \textbf{98.85\%} & \textbf{99.55\%} & \textbf{98.95\%} \\
\hline
  & Exact-match       & 92.31\%  & 91.82\% & 96.63\% \\
2 & Over-provis.      & 98.15\%  & 97.21\% & 98.34\% \\
  & Top-1             & 98.64\%  & 98.07\% & 98.02\% \\ 
\bottomrule
\end{tabular}
\caption{Prediction accuracy of the first layer on three models. Note that the first layer is the hardest one in related works. Other layers have similar or better accuracy than the first layer.}
\label{tab:final-prediction}
\end{table}

\subsection{Expert Loading Performance Analysis}

To establish the practical requirements for expert prediction accuracy, we conducted comprehensive timing benchmarks on representative hardware configurations using DeepSeek-V2-Lite. Our experimental setup measured expert loading latencies across three scenarios: Tesla V100-32GB, A100-40GB, and A100-80GB systems, representing typical deployment environments for MoE inference. Each expert contains 16.5MB of parameters (5.78M parameters × 2 bytes per bfloat16 value), and standard token processing requires loading 6 experts. We measured both disk-to-GPU and memory-to-GPU transfer times using optimized implementations with pinned memory, contiguous tensor layouts, and non-blocking CUDA streams.

\begin{table}[!t]
\centering
\small
\begin{tabular}{lccc}
\toprule
\textbf{Hardware} & \textbf{Disk→GPU} & \textbf{Memory→GPU} \\
\midrule
Tesla V100-32GB & 48.1 ms & 9.5 ms\\
A100-40GB & 49.8 ms& 8.5 ms\\
A100-80GB & 33.5 ms& 4.0 ms\\
\bottomrule
\end{tabular}
\caption{Expert Loading Performance (6 experts, 99MB total)}
\label{tab:expert_loading}
\end{table}

Storage-to-GPU transfers represent the critical bottleneck in MoE inference. As the profiling results in Table \ref{tab:expert_loading} shows, loading 6 experts requires 33.5-49.8ms across the tested hardware, with per-expert costs ranging from 5.6-8.3ms. The high latency is due to storage bandwidth limitations and the inherent serialization of disk I/O operations. Pre-cached experts in system memory achieve dramatically improved transfer rates. Loading 6 experts from memory requires 4.0-9.5ms depending on hardware generation, representing 5-8.4× speedup over disk access. Per-expert memory transfer costs range from 0.7-1.6ms, approaching the theoretical limits imposed by PCIe bandwidth and memory hierarchy overhead.

Analysis of the inference pipeline reveals the computational context for expert prediction and the opportunities for parallel execution. The transformer layer processes tokens through the sequence: pre-attention norm → self-attention → post-attention norm → expert selection + gating → expert computation, illustrated by Figure~\ref{fig:moe_pipeline}. Our prediction system utilizes the weights available immediately after pre-attention norm (0.075-0.129ms) to predict expert requirements in 0.15ms, then executes this prediction pipeline in parallel with the subsequent self-attention (0.738-1.128ms) and post-attention norm (0.080-0.129ms) computations.




\begin{figure}[t]
    \centering     
    
    \begin{subfigure}[b]{\columnwidth}
        \centering 
        \includegraphics[width=\textwidth]{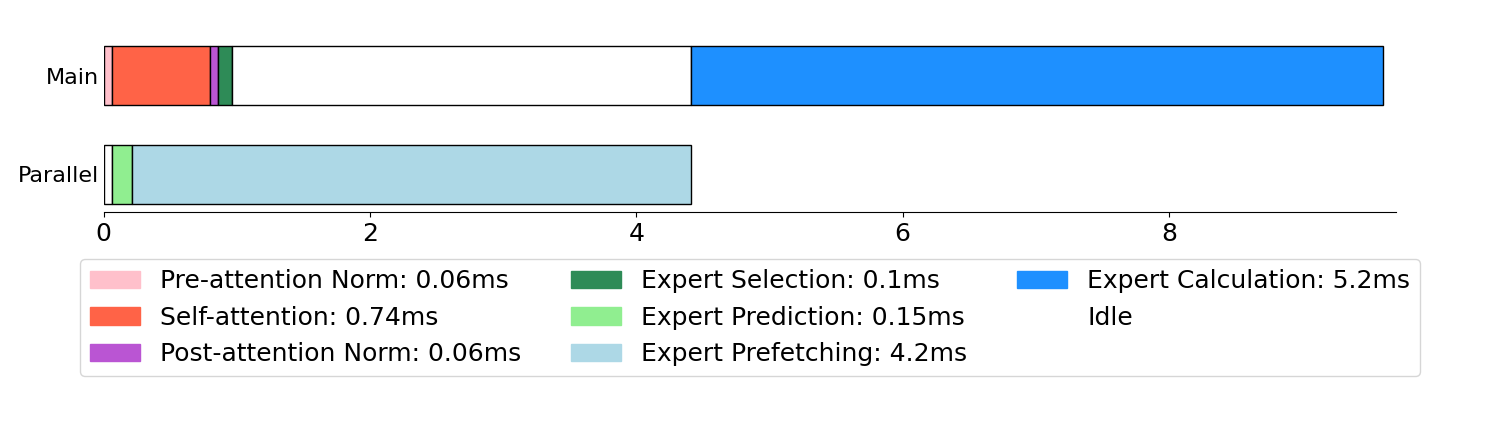}
        \subcaption{A Naive Expert Prefetching Pipeline}  
        \label{fig:prefetch_pipeline}
    \end{subfigure}
    \hfill 
     
    \begin{subfigure}[b]{\columnwidth}
        \centering
        \includegraphics[width=\textwidth]{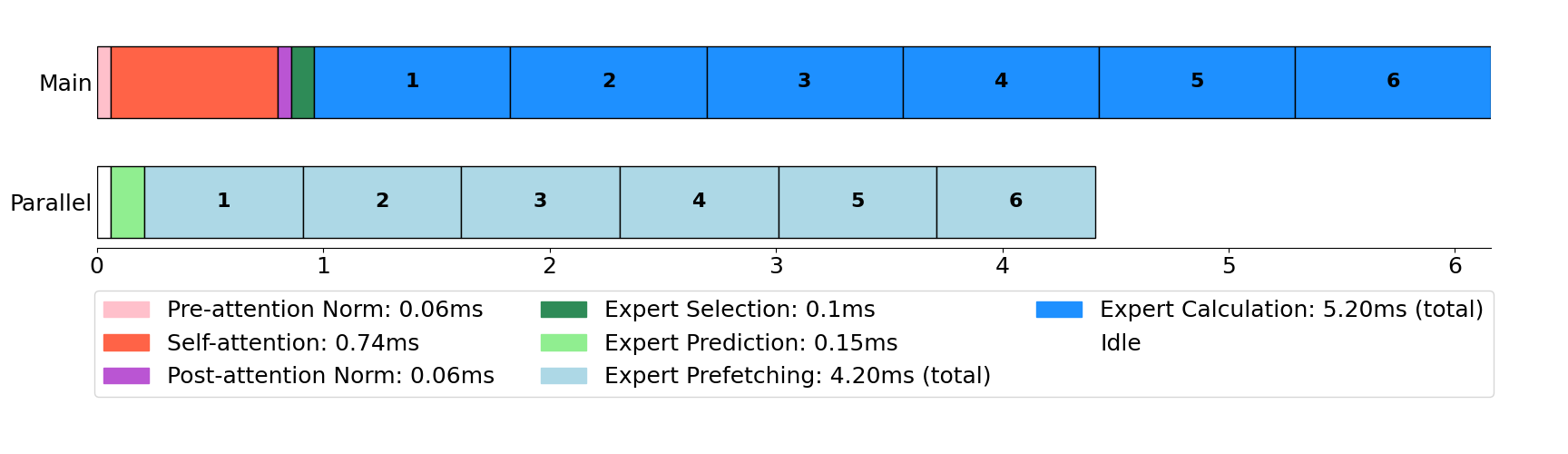}
        \subcaption{Best-Case Scenario with accurate expert prediction}
        \label{fig:pipeline_bc}
    \end{subfigure}
    \hfill 
     
    \begin{subfigure}[b]{\columnwidth}
        \centering
        \includegraphics[width=\textwidth]{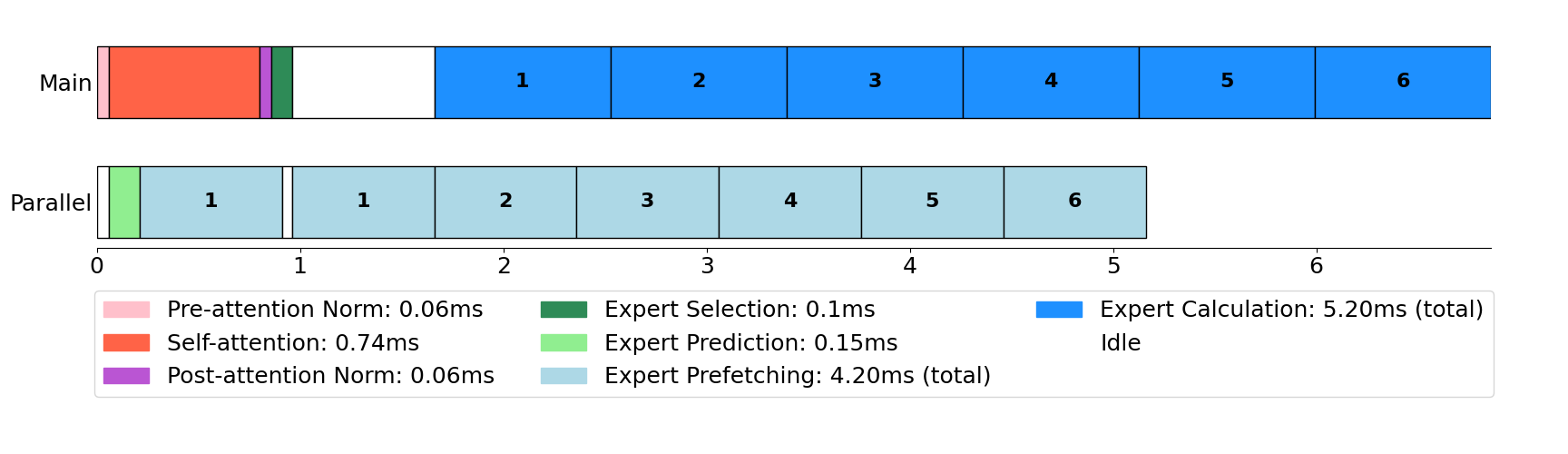}
        \subcaption{Worse-Case Scenario with incorrect expert prediction}
        \label{fig:pipeline_wc}
    \end{subfigure}
    
    \caption{Execution pipelines of a Transformer layer.}
    \label{fig:three_in_one_col} 
\end{figure}

Parallel execution of expert predictor and attention blocks is necessary to leave enough time to prefetch the experts. Compared with the naive parallel execution in Fig. \ref{fig:prefetch_pipeline}, we can apply fine-grained expert computation like Fig. \ref{fig:pipeline_bc} to hide all expert loading latency. This parallel execution strategy provides 0.818-1.257ms of available time for expert prefetching operations before the pipeline reaches the standard expert gating step (0.097-0.143ms). During this window, correctly predicted experts can be loaded from memory (0.7-1.6ms per expert) and made ready for immediate use. When the pipeline reaches expert selection, our predictions are validated against the standard gating decisions. Correctly predicted experts proceed immediately to computation, while mispredicted experts trigger emergency loading (5.6-8.3ms per expert) during the expert computation phase (6.197-10.308ms total). The parallel execution model ensures that prediction accuracy directly impacts overall throughput by enabling immediate expert access, depicted in Figure~\ref{fig:pipeline_bc}, or not requiring additional latency in the event of a misprediction, as shown in Figure~\ref{fig:pipeline_wc}.

\section{Performance Analysis}

\subsection{Accuracy vs. System Complexity}

Our same-layer prediction approach achieves substantial accuracy improvements while reducing system complexity compared to cross-layer methods. The $15\text{-}19$ percentage point improvement over Fate~\cite{FATE_ArXiv_2025} ($93.03\%$ vs $78.79\%$ for DeepSeek-V2) demonstrates that leveraging same-layer information provides superior prediction signals compared to cross-layer approaches. Compared with DuoServe-MoE \cite{DuoServe-MoE_ArXiv_2025} that achieves 54-67\% top-2 accuracy on Mixtral-8x7B and Mixtral-8x22B, our method generalizes to three MoE models with a much higher 93-98\% (+30\%) exact-match accuracy. Compared with caching based method HOBBIT \cite{HOBBIT_ArXiv_2024} with around 55\% hit rate on Mixtral-8x7B and Phi-MoE, we achieve about 40\% improvement in absolute accuracy.

The elimination of cross-layer communication overhead represents a significant system simplification. While Fate requires coordinating information across transformer layers and managing temporal dependencies between layer executions, our approach operates independently within each layer using readily available pre-attention activation tensors.

Prediction latency analysis shows minimal overhead with $0.15$ ms for expert prediction across all model configurations. This overhead represents less than 10\% of the pre-MLP computation pipeline (pre-attention norm + self-attention + post-attention norm), ensuring that the prediction computation remains sufficiently fast to enable parallel fetching of experts while the current layer executes, which preserves the parallelization benefits of expert prediction.

\subsection{I/O Savings Analysis}

The substantial accuracy improvements translate directly into quantifiable I/O performance benefits. Our improved prediction accuracy directly reduces the frequency of expert loading operations during inference. When predictions are correct ($93.03\%$ of tokens), experts are prefetched during self-attention computation, achieving zero loading latency when the pipeline reaches expert selection. When predictions fail ($6.97\%$ of tokens), the system loads experts normally without additional penalty.

The expected expert loading time per token is calculated as the misprediction rate multiplied by the expert loading time. Our approach achieves expected loading times of $0.27\text{-}0.64$ ms/token compared to Fate's $0.85\text{-}2.01$ ms/token across different hardware configurations. On Tesla V100 systems, our expected time is $(1 - 0.9303) \times 9.5 = 0.66$ ms/token compared to Fate's $(1 - 0.7879) \times 9.5 = 2.01$ ms/token, providing $1.37$ ms savings per token.

The frequency-based improvements are substantial: $93.03\%$ of tokens experience zero expert loading latency compared to $78.79\%$ with Fate's approach, representing $14.24$ percentage points more tokens with immediate expert access. Over extended inference sessions of $1000$ tokens, this translates to $569\text{-}1352$ ms total latency savings across different hardware configurations, excluding the reloading time on wrong predictions.

Overprovisioning strategies further improve performance by loading $10$ experts instead of the required $6$, achieving $98.65\%$ prediction accuracy. This reduces expected loading time to $(1 - 0.9865) \times \text{loading time} = 1.35\% \times \text{loading time}$, making $98.65\%$ of tokens achieve zero loading latency at the cost of $67\%$ additional I/O overhead during prefetching.

\subsection{Deployment Strategy Recommendations}

Cloud environment with enough resources should adopt overprovisioning strategies that load $10$ experts instead of required $6$ to achieve $98+\%$ hit rates. The $67\%$ I/O overhead increase might be accommodated by cloud-scale I/O bandwidth~\cite{bodner2025} while the $<2\%$ miss rate minimizes performance-critical cache miss penalties.

Edge devices with I/O bandwidth constraints should utilize strategies based on their parallel loading capacity. For devices that can load multiple experts in parallel with attention computation, precise prediction with $93.03\%$ accuracy provides optimal resource utilization. However, for edge scenarios where I/O bandwidth limits parallel loading to a single expert during the attention computation window, top-1 prediction becomes critical. Our top-1 prediction achieves $98.6\text{-}99.1\%$ hit rates across the different models, ensuring that loading the highest-confidence predicted expert in parallel with attention processing maximizes the utility of the limited parallel execution window.


\section{Discussion}

\subsection{Why the Method Works}

The effectiveness of pre-attention prediction stems from the fundamental information flow within transformer architectures~\cite{abnar2020}. Pre-attention activation tensors represent the token state immediately before expert routing decisions, providing the more relevant and temporally accurate information for predicting routing outcomes.

Unlike cross-layer approaches that attempt to extrapolate routing decisions from previous layer states, our method leverages the input closer to the original routing function. Using the tensor in the same layer eliminates the uncertainty inherent in temporal prediction across layer boundaries and provides access to the complete information set used by the routing mechanism.

The superior performance on simpler models like Phi-mini ($97.62\%$ accuracy) compared to more complex models ($93.03\text{-}94.69\%$ accuracy) might reflect the relationship between routing complexity and prediction difficulty. Models with more specialized expert functions and complex routing patterns present greater prediction challenges, but still benefit substantially from the pre-attention information.

Analysis of model-specific patterns reveals that pre-attention activation tensors capture both semantic content and structural patterns that determine expert routing. The activation tensors encode information about token types, positional patterns, and contextual relationships that routing functions use to make expert selection decisions, consistent with recent studies of attention mechanisms in MoE models~\cite{pikos2025,yang2025umoe}. 

\subsection{Future Work}

End-to-end system integration building on recent advances in MoE system design represents the most immediate opportunity for extending this work. Combining our prediction approach with optimized caching strategies and dynamic overprovisioning based on inference patterns could further improve performance across diverse deployment scenarios. 

Extension to larger MoE models with more experts and more complex routing patterns would validate the approach's scalability. While current results demonstrate effectiveness across models with 2-8 selected experts, larger models with 16+ expert selections might present additional challenges and optimization opportunities.

Dynamic over-provisioning strategies that adjust expert loading based on prediction confidence scores may further optimize the trade-off between I/O overhead and hit rates. Predictions with high confidence scores could operate in exact match mode while less confident predictions could increase over-provisioning to maintain hit rate targets.

\section{Conclusion}

In this paper, we target the expert prediction and prefetching problem in MoE LLMs, especially the limitation of cross-layer prediction on the first few layers. We propose a novel same-layer expert prediction using pre-attention activation tensors to solve this problem. Our key observation is that pre-attention activation tensors contains more recent information than the tensors from the previous layer, and matching the ranking of expert selection scores using simple linear functions is possible. Therefore, we propose two lightweight expert predictors with ranking-aware loss functions, eliminating the architectural complexity and cross-layer communication overhead that limits current approaches. 

Our proposed pre-attention expert prediction achieves $93.03\%$ exact-match accuracy on DeepSeek V2 Lite, $94.69\%$ on Qwen3, and $97.62\%$ on Phi-mini, showing substantial $15-19\%$ absolute improvements over existing cross-layer prediction method FATE \cite{FATE_ArXiv_2025} and $30-40\%$ improvement compared other prediction and caching methods \cite{DuoServe-MoE_ArXiv_2025, HOBBIT_ArXiv_2024}. 

Future MoE architectures can build upon these findings to incorporate native expert prefetching capabilities, while system designers can leverage our parallel execution strategy to optimize inference pipelines across different hardware configurations. Combining high prediction accuracy and reduced system complexity can create more opportunities for MoE system design and deployment strategies.

\bibliography{main}
\bibliographystyle{mlsys2025}



\end{document}